\documentclass[runningheads]{llncs}

\usepackage{eccv}

\usepackage{eccvabbrv}

\usepackage{graphicx}
\usepackage{booktabs}

\usepackage[accsupp]{axessibility}  

\usepackage{hyperref}

\usepackage{orcidlink}

\usepackage{amsmath}
\usepackage{amssymb}
\usepackage{mathtools}
\usepackage{booktabs}
\usepackage{multirow}
\usepackage{colortbl}
\usepackage{xcolor}
\usepackage{pifont}
\usepackage{subcaption}
\usepackage[most]{tcolorbox}
\usepackage{varwidth}
\usepackage{wrapfig}
\usepackage[table]{xcolor}

\definecolor{blockblue}{RGB}{230,245,255}  
\definecolor{blockgreen}{RGB}{235,250,230} 
\definecolor{blockyellow}{RGB}{255,250,220} 

\begin{document}

\title{ProMSA:Progressive Multimodal Search Agents for Knowledge-Based Visual Question Answering} 

\titlerunning{ProMSA:Progressive Multimodal Search Agents for Knowledge-Based Visual Question Answering}

\author{ZhengXian Wu\inst{2,1}\thanks{Equal contribution.} \and
Hangrui Xu\inst{2}$^{\star}$\and
Kai Shi\inst{1}$^{\star, \dagger}$ \and
Zhuohong Chen\inst{2} \and
Yunyao Yu\inst{2} \and
Chuanrui Zhang\inst{3} \and
Zirui Liao\inst{2} \and
Jun Yang\inst{1}$^{\star\star}$ \and
Zhenyu Yang\inst{1} \and
Haonan Lu\inst{1} \and
Haoqian Wang\inst{2}\thanks{Corresponding author.}}

\authorrunning{Z.~Wu et al.}

\institute{OPPO AI Center, OPPO Inc. China \and
The Shenzhen International Graduate School, Tsinghua University
 \and Nanyang Technological University, Singapore\\
\email{zx-wu24@mails.tsinghua.edu.cn}}

\maketitle

\begingroup
\renewcommand{\thefootnote}{\ensuremath{\dagger}}
\footnotetext{Project leader.}
\endgroup

\begin{abstract}
    Knowledge-based Visual Question Answering (KB-VQA) requires models to combine image understanding with external knowledge. Most prior methods use a fixed retrieve-then-generate pipeline with a pre-selected retriever and a static top-k setting, which is not adaptive during reasoning. We propose \textbf{ProMSA}, a progressive multimodal search agent for KB-VQA. Given an image-question pair, the agent iteratively chooses image search, text search, or stop, under explicit tool-call budgets and with deduplication to avoid redundant retrieval. For training, we first use rejection-sampling SFT to learn valid tool-use formats, then optimize the agent with \textbf{TN-GSPO}, a sequence-level RL objective that normalizes updates by both generation length and tool-interaction depth. Experiments on E-VQA and InfoSeek show consistent gains over strong RAG and agent baselines, and improved retrieval and end-to-end accuracy. The code is available at https://github.com/DingWu1021/Promsa.
  \keywords{Knowledge-based Visual Question Answering \and Reinforcement Learning \and Multimodal Search Agent}
\end{abstract}

\section{Introduction}
\label{sec:intro}

In recent years, multimodal large language models (MLLMs) have achieved strong progress on general visual question answering\cite{yin2024survey}. 
With large scale pretraining on paired image and text data, followed by task specific fine tuning, these models have achieved strong performance on visual math reasoning\cite{wang2025mv}, image understanding\cite{chen2025compcap}, and complex scene inference tasks\cite{wang2025escapecraft}.
However, their capability remains limited for knowledge-based visual question answering (KB-VQA) under long-tail entity distributions\cite{caffagni2024wikillava}.
KB-VQA requires the model to recognize entities in the image and to reason with external knowledge sources such as Wikipedia when necessary\cite{yang2025omgm}.
In the long-tail setting, entities are often rare, making it difficult for the model to reliably decide whether the entity has been correctly identified and whether the retrieved evidence is trustworthy.

Early approaches mostly follow a fixed retrieval augmented generation pipeline\cite{yan2024echosight}.
They first run a single image retrieval step to collect the top K relevant Wikipedia pages, rerank the retrieved results, and then append the retrieved text to the prompt for answer generation.
These methods suggest a clear pattern: when key facts are retrieved from external sources, models can shift from relying on parametric knowledge to producing answers grounded in retrieved evidence, leading to a significant improvement in accuracy.
Recent work further improves this paradigm along two directions\cite{cocchi2025augmenting}. 
The first direction focuses on retrieval quality by increasing recall and reducing missing evidence.
The second direction focuses on knowledge usage by filtering noisy content and keeping the input context centered on information that supports the answer.
In terms of specific designs, Wiki-PRF\cite{hong2025knowledge} introduces extra processing before retrieval, such as generating a summary and cropping the image, and then performs more detailed retrieval with both image and text retrievers. 
ReAG\cite{compagnoni2025reag} trains a critic model to judge the relevance of candidate passages and explicitly removes irrelevant content, preventing noisy evidence from being directly fed into the generator.

\begin{figure}[tb]
  \centering
  \includegraphics[height=6.3cm]{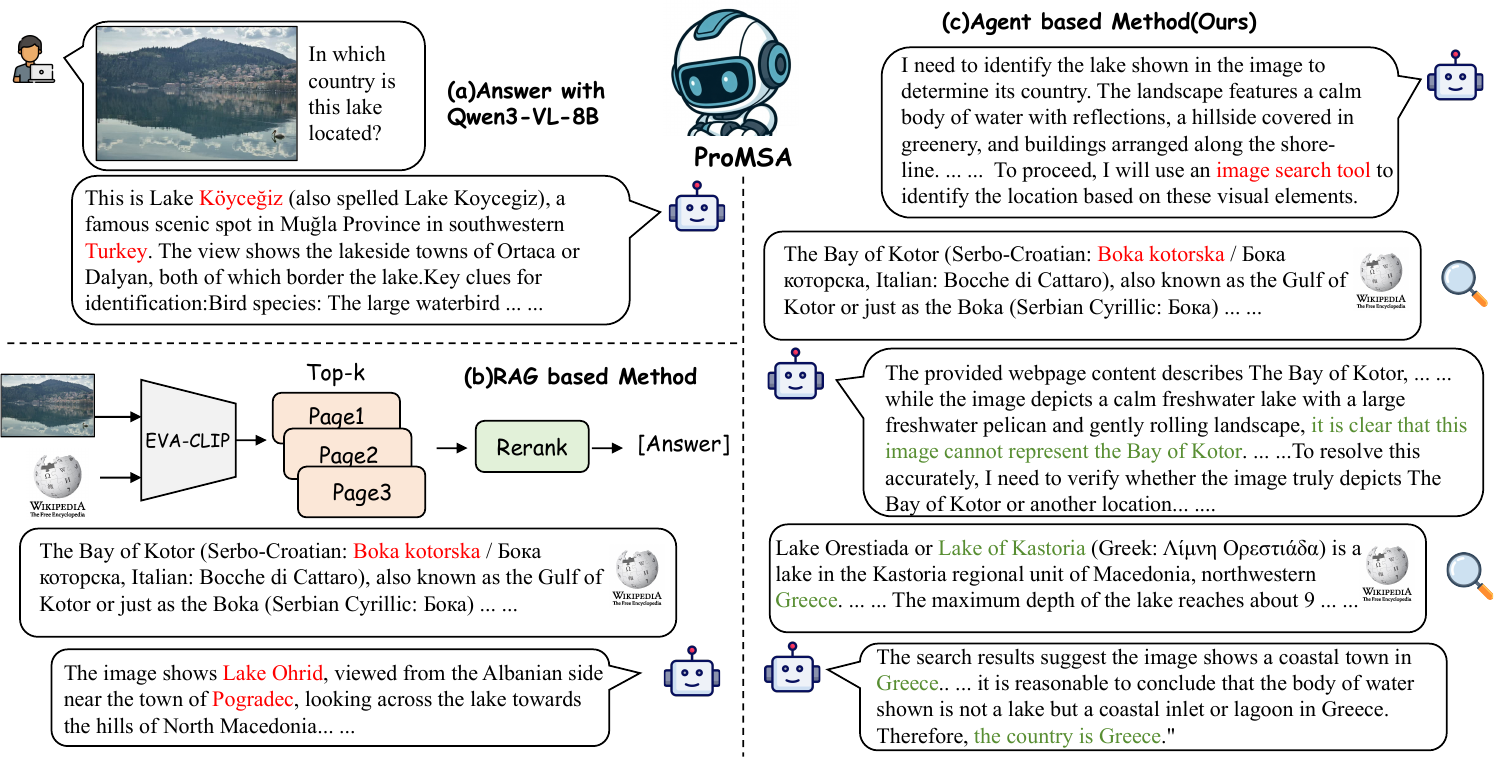}
  \caption{Comparison between direct answering, RAG-based retrieval, and our progressive multimodal search agent for KB-VQA.}
  \label{fig:fig1}
\end{figure}

However, such fixed pipelines still have limitations for KB-VQA.
\textbf{(1)Tools and retrieval policies are not adaptive. }
Existing methods either manually select retrievers for each dataset or combine image and text retrieval to increase recall.
In KB-VQA, the problem state is highly diverse. 
Some can be answered without retrieval, some require identifying the entity first, and others involve known entities but missing attributes.
\textbf{(2) There is no mechanism to recover from failures. }
When retrieved results are unreliable or evidence is insufficient, a fixed pipeline still forces the model to enter the answer stage. 
The model cannot reflect on the evidence and correct it through query rewriting or additional retrieval, which often leads to confident answers based on incorrect evidence.
\textbf{(3) It is hard to support multi-hop evidence chains. }
KB-VQA often requires filling in facts step by step. 
Static top K evidence injection cannot progressively expand the evidence required for reasoning and therefore often fails on multi-hop questions.
These issues suggest that KB-VQA does not mainly need a larger top K or a stronger reranker. 
Instead, it needs a closed loop system that tightly couples retrieval and reasoning. 
The model should behave like a budget aware researcher. 
It retrieves when uncertain, continues searching when evidence is insufficient, and stops once the evidence is enough.
Recent work\cite{wu2025mmsearch} has explored search-agent approaches for general multimodal retrieval.
However, most of these agents are designed mainly for multi-hop tasks, rather than for adaptive retrieval.
Moreover, they often lack an explicit correction mechanism: once the first retrieval drifts to a wrong entity, later interactions tend to accumulate evidence around incorrect pages, leading to a bias that is difficult to correct.

Based on these insights, we propose \textbf{ProMSA}, a progressive multimodal search agent for KB-VQA(Fig.\ref{fig:fig1}).
The agent enables the model to adaptively switch between image retrieval and text retrieval and to gather evidence over multiple rounds for reasoning.
Since each tool call consumes time and compute budget, we reformulate KB-VQA as a budgeted problem of progressive search and reasoning. 
At each step, the model decides whether to perform image retrieval, perform text retrieval, or stop and answer based on the currently observed evidence and the history of previous steps.
Unlike fixed pipelines, our agent supports multiple tool calls with explicit deduplication. 
When entity identification remains uncertain, the model can trigger image retrieval again while excluding previously retrieved candidates to move toward the correct page under appearance variations.
When the entity is identified but the required knowledge is still missing, the model can rewrite the query and invoke text retrieval to fill in missing attributes.
This design enables adaptive retrieval, allowing the model to select the appropriate tool at different stages of the reasoning process.
At training time, we first cold-start the model with a small SFT set built by rejection sampling, so that it learns basic tool-call formats and interaction patterns. 
We then apply reinforcement learning to further improve the search policy, enabling the model to learn more effective tool usage over multiple interaction rounds. 
Specifically, we propose TN-GSPO, which extends sequence-level normalization from only generation length to also account for tool interaction depth, aligning the update scale with the decision structure of budgeted search and leading to more stable search strategies.
Experiments on multiple benchmarks show that ProMSA consistently outperforms strong baselines and achieves state-of-the-art performance on KB-VQA tasks.Our contributions are summarized as follows:
\begin{itemize}
\item We propose \textbf{ProMSA}, a progressive multimodal search agent for KB-VQA that couples retrieval and reasoning through multi-round interactions.
\item We formulate KB-VQA as a \textbf{budgeted progressive search-and-reasoning problem} that learns when to retrieve, which modality to use, and when to stop.
\item We introduce \textbf{TN-GSPO}, a tool-horizon normalized sequence-level policy optimization method for stable learning of search policies.
\item Experiments on multiple KB-VQA benchmarks show that ProMSA consistently outperforms strong baselines and achieves state-of-the-art performance.
\end{itemize}

\begin{figure}[tb]
  \centering
  \includegraphics[height=7.0cm]{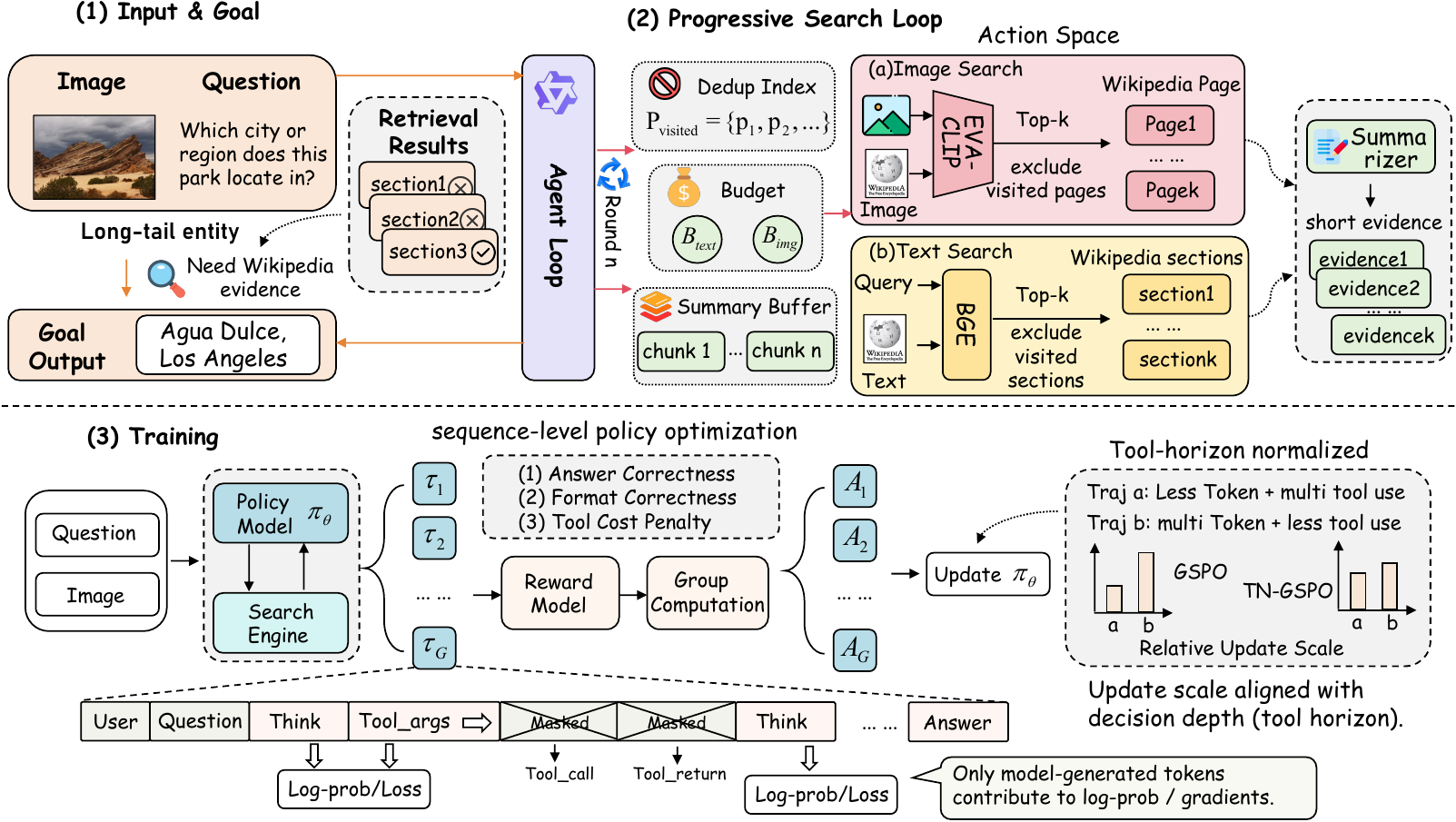}
  \caption{Overview of our progressive multimodal search agent for KB-VQA.
Given an image-question pair, the agent iteratively performs image and text search over Wikipedia, and is trained with tool-horizon normalized sequence-level RL (TN-GSPO).}
  \label{fig:baseline}
\end{figure}

\section{Related Work}
\subsection{Knowledge-Based Visual Question Answering}
Knowledge-based visual question answering (KB-VQA) extends conventional VQA by requiring models to combine visual understanding with external knowledge sources. 
A major line of work focuses on improving multimodal retrieval and evidence construction. 
OMGM\cite{yang2025omgm} proposes a coarse-to-fine framework that coordinates multiple modalities and granularities through hierarchical reranking and section-level evidence selection. 
Another line of work targets knowledge filtering and relevance estimation after retrieval. 
QKVQA\cite{ye2026qkvqaquestionfocusedfilteringknowledgebased} identifies limitations in both cross-document selection and intra-document localization, and proposes question-aware filtering with dynamic multi-article selection. 
More recent work explores consistency between parametric knowledge and retrieved evidence. 
CC-VQA\cite{hong2026cc} studies conflicts between internal model knowledge and external contexts and introduces conflict-aware reasoning to mitigate inconsistency. 

Overall, existing KB-VQA methods have substantially improved retrieval precision, evidence filtering, and conflict mitigation, but most still operate within a largely static retrieval-then-generate pipeline. 

\subsection{Search-Based Agents}
Recent advances in large reasoning models have enabled search-based agents that can call external tools, issue search queries during reasoning, and iteratively incorporate retrieved evidence for multi-step inference.
Compared with conventional one-shot RAG, this line of work emphasizes adaptive tool usage, sequential evidence acquisition, and trajectory-level optimization. 

In the language domain, Search-o1\cite{li2025searcho1} integrates search into the reasoning loop and introduces a Reason-in-Documents module to refine retrieved content before incorporating it into the reasoning process. 
Search-R1\cite{jin2025searchr1} further formulates search-augmented reasoning as a reinforcement learning problem, enabling models to generate search queries during intermediate reasoning steps.
This paradigm has recently been extended to multimodal agentic search. 
MMSearch-R1\cite{wu2025mmsearch} introduces an RL framework for on-demand multi-turn search with both image and text retrieval tools. 
SenseNova-MARS\cite{chng2025sensenova} further generalizes this idea by integrating multiple tools, including image search, text search, and visual cropping, within a unified multimodal reasoning-and-search framework.

These search-agent systems provide a strong foundation for adaptive retrieval, but they are primarily developed for general web search or open-domain reasoning. 
In contrast, KB-VQA requires retrieving knowledge about visual entities in the image, which is often uncertain or long-tailed and cannot be reliably solved by a single retrieval step.

\section{Method}
\subsection{Progressive Retrieval--Reasoning Pipeline}
The core challenge of KB-VQA is that the model must decide how to acquire external knowledge during reasoning.
In practice, the model faces two major sources of uncertainty.
\textbf{(1) Retrieval modality uncertainty.}
When the entity in the image cannot be identified, the model should perform image retrieval to locate the corresponding Wikipedia page.
When the entity is already known but the question requires attribute completion or multi-hop reasoning, the model should generate a high-quality text query and perform text retrieval to gather missing evidence.
\textbf{(2) Retrieval depth uncertainty.}
Because the user image and knowledge-base images may differ in viewpoint, background, and context, the correct entity may appear as the top candidate, or it may require multiple retrieval attempts with explicit de-duplication to gradually reach the right page.

Motivated by these observations, we aim to learn an adaptive strategy that allows the model to autonomously select the retrieval modality and depth(Fig.\ref{fig:baseline}).
Specifically, we formulate KB-VQA as a budgeted progressive retrieval--reasoning process: under a limited number of tool calls, the model alternates between retrieval and reasoning, accumulates information relevant to the question, and stops to answer once the collected evidence is sufficient.

Given an input image and question $x = (I, q)$, the goal is to produce the final answer $a$.
Unlike RAG, we allow multi-turn interactions.
At step $t$, the model selects an action $u_t$ based on the information state $s_t$.
The action space is
\begin{equation}
u_t \in \mathcal{U} = \{\texttt{img\_search},\ \texttt{text\_search},\ \texttt{stop}\}.
\end{equation}
Each action produces an external observation $o_{t+1}$ (e.g., retrieved content), and the internal state is updated by $s_{t+1} = \text{Update}(s_t, o_{t+1})$. We impose budget constraints on tool usage. Let $B_{\text{img}}, B_{\text{text}}$ denote the maximum number of calls to image search and text search, respectively.
Then any trajectory $\tau$ must satisfy
\begin{equation}
\left| \{ t \mid u_t = \texttt{img\_search} \} \right| \le B_{\text{img}}, 
\qquad
\left| \{ t \mid u_t = \texttt{text\_search} \} \right| \le B_{\text{text}}.
\end{equation}
When the model outputs \texttt{stop}, it enters the final answering stage and generates the answer $a$.

\subsection{Progressive Multimodal Search-Agent}
We parameterize the policy with a single multimodal large model, denoted as $\pi_\theta$.
At each step, the policy generates a structured output conditioned on the current state $s_t$:
\begin{equation}
y_t = (z_t,\, u_t,\, \text{args}_t)
\sim \pi_\theta(\cdot \mid s_t).
\end{equation}
Here, $z_t$ is the model-generated reasoning,
$u_t$ is an action token (\texttt{img\_search}, \texttt{text\_search}, or \texttt{stop}),
and $\text{args}_t$ specifies action arguments such as query rewrites, top-$k$, and the de-duplication list.
When $u_t=\texttt{stop}$, the model enters the final answering stage and generates an answer:
\begin{equation}
a \sim \pi_\theta(\cdot \mid s_t, u_t=\texttt{stop}).
\end{equation}

Using a unified generative policy allows retrieval decisions and reasoning to share the same semantic representation.
This enables the agent to jointly handle both entity alignment and knowledge completion within a single trajectory.
In contrast to fixed pipelines that decouple retrieval from reasoning, our end-to-end policy learns action selection and stopping directly from interaction trajectories, and allocates retrieval steps more effectively under a limited budget.

We provide two external tools for retrieval. The image retrieval tool takes the input image and an exclusion list, while the
text retrieval tool takes a query generated by the model.
Both tools return candidate content:
\begin{equation}
\texttt{img\_search}(I,\text{exclude}) \rightarrow \{p_i\}_{i=1}^{K_{\text{img}}},
\quad
\texttt{text\_search}(q_t,\text{exclude}) \rightarrow \{p_i\}_{i=1}^{K_{\text{text}}}.
\end{equation}
The query $q_t$ is produced by the model through $\text{args}_t$ and is used to
retrieve additional evidence when the entity has been mostly identified.

During inference, we maintain the set of retrieved pages $\mathcal{P}_t$.
At each tool call, we pass it as the exclusion list and update it after receiving new candidates:
\begin{equation}
\text{exclude}_t = \mathcal{P}_t, \quad
\mathcal{P}_{t+1} = \mathcal{P}_t \cup \{p_i\}.
\end{equation}

Since \texttt{img\_search} returns an entire Wikipedia page and \texttt{text\_search} may also return long text chunks, directly injecting the raw content would cause severe context growth.
We therefore compress the returned content into short, question-relevant summaries, and use them as observations for the next decision step.
Let $p$ denote a retrieved page and $W(p)$ its full page content.
We summarize the retrieved content into a question-conditioned snippet:
\begin{equation}
\hat{w}(p) = \text{Summarize}(q, W(p)).
\end{equation}
The summary $\hat{w}(p)$ is a short text that preserves the most relevant entity description, key attributes, and relation cues for answering the question.
With this design, the observation returned by a tool call can be written in a unified form:
$o_{t+1} = \{\hat{s}_{t+1}^{(j)}\}_{j=1}^{m_{t+1}}$,
where each snippet $\hat{s}_{t+1}^{(j)}$ is a summarized text segment derived from either image retrieval or text retrieval.
The agent appends these snippets to the context to form the next state
$s_{t+1} = \text{Update}(s_t, o_{t+1})$.

\subsection{Training}
\subsubsection{Rejection-Sampling SFT.}
In the Search-Agent setting, the model must first learn how to issue valid tool calls, including the required format and argument structure.
Directly applying RL from scratch is often unstable: the exploration space is large, and early trajectories frequently receive zero reward due to malformed calls or invalid executions.
Therefore, we warm-start the policy with a small cold-start dataset constructed via rejection sampling, so that the policy becomes executable before RL.

For each sample $x=(I,q)$, we sample $N$ trajectories from an initial policy $\pi_{\theta_0}$:
\begin{equation}
\tau_i \sim \pi_{\theta_0}(\cdot \mid x),\quad i=1,\dots,N.
\end{equation}
Each trajectory contains multiple rounds of model outputs and tool interactions:
\begin{equation}
\tau = (y_1, o_2, y_2, o_3, \dots, y_T),
\end{equation}
where $y_t$ is the model-generated output at step $t$ (e.g., an action, its arguments, or the final answer), and $o_{t+1}$ is the tool response.

We apply rule-based filters and keep only trajectories that are suitable for SFT, forming $\mathcal{D}_{\text{sft}}$:
(i) the tool-call format is valid (all required JSON fields are present and the budget is not exceeded);
(ii) the trajectory is executable (the tool returns a response);
(iii) the final answer is correct.
We retain at most one valid trajectory per question for SFT training.
We then perform standard maximum-likelihood training:
\begin{equation}
\mathcal{L}_{\text{SFT}}(\theta)=
-\mathbb{E}_{(x,\tau)\sim \mathcal{D}_{\text{sft}}}
\sum_{t\in T_{\text{gen}}}\log \pi_\theta\!\left(y_t \mid y_{<t}, s_t\right).
\end{equation}

\subsubsection{Reinforcement Learning.}
In the KB-VQA Search-Agent setting, the trajectory return is mainly determined by whether the final answer is correct, which typically can only be verified after multiple rounds of tool interaction. 
Compared with dense token-level rewards, our agent is closer to a small number of discrete decisions that determine success or failure. 
We therefore adopt a GSPO-style \emph{sequence-level} policy optimization method, where the update is driven by the return of the whole trajectory.

Given an input $x=(I,q)$, the policy $\pi_\theta$ samples a trajectory.
We partition tokens into a set of trainable generated tokens $T_{\text{gen}}$ and a set of tool-returned tokens $T_{\text{tool}}$.
All tokens in $T_{\text{tool}}$ are masked out, so the trajectory log-probability is defined only on $T_{\text{gen}}$:
\begin{equation}
\log \pi_\theta(\tau)=\sum_{t\in T_{\text{gen}}}\log \pi_\theta(y_t\mid s_t),
\end{equation}
where $s_t$ includes the history of generated tokens and tool observations, but only generated tokens contribute to gradients.

\paragraph{Reward.}
We use a sparse sequence-level return aligned with the search budget:
\begin{equation}
R(\tau)= r_{\text{ans}}(\tau)+r_{\text{fmt}}(\tau)-\lambda\,C(\tau), \quad
C(\tau)=\frac{\#\text{tool\_calls}(\tau)}{H_{\max}} .
\end{equation}
where $r_{\text{ans}}$ measures answer correctness, $r_{\text{fmt}}$ measures format compliance, and $C(\tau)$ is a tool-cost term that encourages efficient retrieval under a fixed budget. 
$H_{\max}$ is the maximum number of allowed tool calls.

\paragraph{Tool-Normalized GSPO .}
Standard GSPO normalizes by the generation length to stabilize update magnitudes. 
However, in our setting, trajectory difficulty is mainly driven by the \emph{external interaction depth}. 
For trajectories with similar tool usage, variation in text length can cause large fluctuations in $|T_{\text{gen}}|$, which introduces a length bias (e.g., preferring shorter outputs to obtain a more stable optimization signal).
To address this mismatch, we incorporate the decision depth into the normalization of the policy ratio, leading to \textbf{Tool-Normalized GSPO (TN-GSPO)}.
Let $\Delta_t=\log\pi_\theta(y_t\mid s_t)-\log\pi_{\text{ref}}(y_t\mid s_t)$, and define
\begin{equation}
L(\tau)=|T_{\text{gen}}|,\quad
H(\tau)=1+\#\text{tool\_calls}(\tau),\quad
D(\tau)=L(\tau)\big(1+c\,H(\tau)^{\alpha}\big).
\end{equation}
where $L(\tau)$ counts only trainable generated tokens and $H(\tau)$ counts only the tool-interaction depth.
We then define the sequence-level likelihood ratio as
\begin{equation}
r_\theta(\tau)=\exp\!\Big(\frac{1}{D(\tau)}\sum_{t\in T_{\text{gen}}}\Delta_t\Big).
\end{equation}
and use a centered advantage $A(\tau)=R(\tau)-b(x)$.
The final objective is
\begin{equation}
\mathcal{L}_{\text{TN-GSPO}}(\theta)=
-\mathbb{E}_{\tau\sim\pi_\theta}\!\Big[
\min\Big(
r_\theta(\tau)\,A(\tau),\;
\text{clip}\!\big(r_\theta(\tau),1-\epsilon^{-},1+\epsilon^{+}\big)\,A(\tau)
\Big)
\Big].
\end{equation}
For stable sequence-level optimization, we use asymmetric clipping, where $\epsilon^{-}$ and $\epsilon^{+}$ control clipping in the decreasing and increasing directions, respectively.
This normalization makes the update scale depend on both the number of trainable tokens and the depth of tool interaction, which better matches budgeted search and reduces bias caused by fluctuations in generation length.

\section{Experiments}

\begin{table*}[t]
    \scriptsize
    \caption{Performance comparison on E-VQA and InfoSeek.* indicates that the original SerpApi retrieval service is replaced with a local retrieval system, and † denotes results reproduced with a different backbone model}
    \setlength{\tabcolsep}{3.5pt}
    \label{tab:main_results}
    \hspace{-5mm}
    \begin{tabular}{l l l
                    c c
                    c c c}
    \toprule
    \multicolumn{3}{c}{} &
    \multicolumn{2}{c}{\textbf{E-VQA}} &
    \multicolumn{3}{c}{\textbf{InfoSeek}} \\
    \cmidrule(lr){4-5} \cmidrule(lr){6-8}
    \textbf{Method} &
    \textbf{Retriever} &
    \textbf{Model} &
    \textbf{Single-Hop} & \textbf{All} &
    \textbf{Un-Q} & \textbf{Un-E} & \textbf{All} \\
    \midrule

    \rowcolor{blockgreen}
    \multicolumn{8}{c}{\textbf{\textit{Zero-shot MLLMs}}} \\

    BLIP-2\cite{li2023blip} &
        - &
        Flan-T5 &
        12.6 & 12.4 &
        12.7 & 12.3 & 12.5 \\

    GPT-4V\cite{achiam2023gpt} &
        - &
        - &
        26.9 & 28.1 &
        15.0 & 14.3 & 14.6 \\

    Qwen2.5-VL-7B\cite{bai2025qwen25vltechnicalreport} &
        - &
        - &
        22.1 & 20.6 &
        22.8 & 24.1 & 23.7 \\

    Qwen3-VL-2B\cite{bai2025qwen3} &
        - &
        - &
        22.2 & 21.9 &
        18.6 & 19.2 & 18.9 \\

    Qwen3-VL-8B &
        - &
        - &
        25.3 & 24.8 &
        25.9 & 25.5 & 25.7 \\

    \midrule

    \rowcolor{blockblue}
    \multicolumn{8}{c}{\textbf{\textit{Search-Agent Models}}} \\

    MMSearch-R1*\cite{wu2025mmsearch} &
        BGE+EVA-CLIP &
        Qwen2.5-VL-7B &
        40.6 & 40.7 &
        40.3 & 39.3 & 39.7 \\

    DeepEyesV2*\cite{hong2025deepeyesv2} &
        BGE+EVA-CLIP &
        Qwen2.5-VL-7B &
        40.1 & 39.5 &
        41.6 & 40.8 & 41.3 \\

    \midrule

    \rowcolor{blockyellow}
    \multicolumn{8}{c}{\textbf{\textit{Retrieval-Augmented Models}}} \\

    EchoSight\cite{yan2024echosight} &
        EVA-CLIP-8B &
        Mistral-7B &
        26.4 & 24.9 &
        30.0 & 30.7 & 30.4 \\

    CC-VQA\cite{hong2026cc} &
        EVA-CLIP-8B &
        Qwen2.5-VL-7B &
        41.4 & 36.1 &
        44.7 & 46.1 & 45.1 \\

    EchoSight$^\dagger$ &
        EVA-CLIP-8B &
        Qwen3-VL-8B &
        38.1 & 35.2 &
        32.6 & 32.2 & 32.4 \\

    REAL\cite{ye2026real} &
        EVA-CLIP-8B &
        Qwen3-VL-8B &
        45.5 & 41.4 &
        43.1 & 45.1 & 44.1 \\

    MaS-VQA\cite{mao2026mas} &
        EVA-CLIP-8B &
        Qwen3-VL-8B &
        42.2 & 41.3 &
        43.7 & 43.9 & 43.8 \\

    \midrule

    \textbf{ProMSA(Ours)} &
        BGE+EVA-CLIP &
        Qwen2.5-VL-7B &
        50.0 & 49.7 &
        48.8 & 49.6 & 49.2 \\

    \textbf{ProMSA(Ours)} &
        BGE+EVA-CLIP &
        Qwen3-VL-2B &
        42.4 & 41.2 &
        44.1 & 43.0 & 43.6 \\

    \textbf{ProMSA(Ours)} &
        BGE+EVA-CLIP &
        Qwen3-VL-8B &
        \textbf{52.2} & \textbf{52.6} &
        \textbf{53.6} & \textbf{53.3} & \textbf{53.4} \\

    \bottomrule
    \end{tabular}
\end{table*}

\subsection{Dataset}
Our training data is built upon the training splits of Encyclopedic-VQA\cite{mensink2023encyclopedic} (E-VQA) and InfoSeek\cite{chen2023can}.

E-VQA contains 221K question–answer pairs covering around 16.7K fine-grained Wikipedia entities. 
Following the reasoning difficulty, questions are grouped into single-hop and multi-hop types. 
Single-hop questions can be answered by retrieving information from a single Wikipedia page, while multi-hop questions require sequential retrieval and reasoning across multiple pages, testing multi-step knowledge integration. 
E-VQA also provides a Wikipedia-based knowledge base with about 2M pages, each containing a title, text sections, and related images.

InfoSeek consists of about 1.3M image–question–answer triplets, associated with around 11K distinct Wikipedia pages. 
It is split into train/validation/test sets with approximately 934K, 73K, and 348K samples, respectively. 
Notably, both the validation and test sets include questions about unseen entities, which poses a stronger generalization challenge under long-tailed knowledge distributions. 
InfoSeek provides an external Wikipedia knowledge base containing around 6M entities. 
Following prior work, we use a 100K-page subset as the retrieval corpus in our experiments to ensure fair comparison with existing methods.
More details about datasets and baselines are provided in the supplementary material.

\begin{table}[t]
\centering
\footnotesize
\setlength{\tabcolsep}{6pt}
\caption{Performance across different training stages.}
\label{tab:training_stages}
\begin{tabular}{l l c c c}
\toprule
\multicolumn{2}{c}{Dataset} & Base & \textbf{Stage I. Cold Start} & \textbf{Stage II. RL} \\
\midrule

\multirow{2}{*}{E-VQA}
& Single-Hop & 33.2 & 38.4 & \textbf{52.2} \\
& All        & 32.8 & 38.6 & \textbf{52.6} \\

\midrule

\multirow{3}{*}{InfoSeek}
& Unseen-Q & 36.8 & 42.7 & \textbf{53.6} \\
& Unseen-E & 36.1 & 41.6 & \textbf{53.3} \\
& All  & 36.4 & 42.1 & \textbf{53.4} \\

\midrule
Average & All & 35.1 & 40.7 & \textbf{53.0} \\
\bottomrule
\end{tabular}
\end{table}

\begin{table*}[t]
\centering

\begin{minipage}{0.48\textwidth}
\centering
\scriptsize
\setlength{\tabcolsep}{6pt}
\caption{Comparison of sequence-level policy optimization methods. \textsuperscript{*} indicates using asymmetric clipping.}
\label{tab:RL_method}
\begin{tabular}{c|cc}
\toprule
\textbf{Method} & \textbf{E-VQA} & \textbf{InfoSeek} \\
\midrule
GRPO\cite{shao2024deepseekmath}  & 44.2 & 43.7 \\
GRPO\textsuperscript{*} & 49.7 & 50.2 \\
GSPO\textsuperscript{*}\cite{zheng2025group}  & 49.3 & 49.6 \\
\rowcolor{green!12}
\textbf{TN-GSPO\textsuperscript{*}} & \textbf{52.6} & \textbf{53.4} \\
\bottomrule
\end{tabular}
\end{minipage}
\hfill
\begin{minipage}{0.48\textwidth}
\centering
\scriptsize
\setlength{\tabcolsep}{3pt}
\caption{Impact of equipping the agent with different tools. The maximum number of interaction rounds follows the original per-tool budget.}
\label{tab:tool_ablation}
\begin{tabular}{cc|cc}
\toprule
\textbf{Text} & \textbf{Image} & \textbf{E-VQA} & \textbf{InfoSeek} \\
\midrule
$\checkmark$ & $\times$ & 27.6 & 36.8 \\
$\times$     & $\checkmark$ & 34.7 & 21.4 \\
\rowcolor{green!12}
$\checkmark$ & $\checkmark$ & \textbf{52.6} & \textbf{53.4} \\
\bottomrule
\end{tabular}
\end{minipage}

\end{table*}



\subsection{Implementation Details}
We train our agent on Qwen2.5-VL-7B\cite{bai2025qwen25vltechnicalreport} and Qwen3-VL-2B/8B\cite{bai2025qwen3} using a two-stage pipeline: supervised fine-tuning (SFT) followed by reinforcement learning (RL). We implement SFT with \textsc{LLaMA-Factory}\cite{zheng2024llamafactory} and RL with \textsc{veRL}\cite{sheng2024hybridflow}.

In the SFT stage, we sample 3{,}000 training instances from the E-VQA and InfoSeek training sets for cold-start training. 
We only fine-tune the language model while freezing the visual encoder and the multimodal projector. 
We train for 3 epochs with a learning rate of $1\times 10^{-5}$.
This stage mainly ensures the policy is executable and can use the tool set in a basic and reliable way.

In the RL stage, we further improve the model's tool usage. 
We sample 15{,}000 instances from the E-VQA and InfoSeek training sets and train for 3 epochs. 
We use a global batch size of 128 and a learning rate of $1\times 10^{-6}$. 
For stable sequence-level optimization, we use asymmetric clipping following DAPO with the Clip-Higher strategy, setting $\texttt{low}=0.2$ and $\texttt{high}=0.28$. 
Each training trajectory allows up to $T=7$ interaction steps. 
We use a tool-cost penalty factor $\lambda=0.5$, and $c=0.04$ , $\alpha=1$.
At each step, the agent can generate up to 4{,}096 tokens, and the total token budget per trajectory is capped at 16{,}384.

For retrieval, we use EVA-CLIP\cite{sun2023eva} and BGE\cite{chen2024m3} as the image and text retrievers, respectively, and deploy Qwen3-VL-8B as the summarization service. 
Both image search and text search can be called at most 3 times. 
For each call, we retrieve Top-3 Wikipedia pages for image search or Top-3 sections for text search. 
We report results following the official evaluation protocols of each dataset. 
For E-VQA, we use BERT-based Matching (BEM) to measure semantic agreement between the generated answer and the reference answer\cite{yan2024echosight}. 
For InfoSeek, the metric depends on the question type\cite{yan2024echosight}; we report both standard VQA Accuracy and Relaxed Accuracy for a comprehensive evaluation. 
All training is conducted on $8\times$ A800 GPUs.

\begin{table*}[t]
\centering
\scriptsize
\setlength{\tabcolsep}{4pt}

\begin{minipage}{0.52\textwidth}
\centering
\caption{Inference time comparison with existing methods (seconds per sample). 
Methods marked with $^*$ use LLaMA-3.1-8B as the generator, while others use Qwen-2.5-VL-7B.}
\label{tab:time_comparison}

\begin{tabular}{c|cc}
\toprule
\textbf{Method} & \textbf{E-VQA (time)} & \textbf{E-VQA (accuracy)} \\
\midrule
Echosight & \textbf{1.2 (1.00x)} & 18.8 \\
\midrule
MMsearchR1 & 1.7 (1.42x) & 40.7 \\
Deepeyesv2 & 2.4 (2.00x) & 39.5 \\
ReflectiVA$^*$\cite{cocchi2025augmenting} & 1.5 (1.25x) & 35.5 \\
\midrule
\textbf{Ours} & 1.8 (1.50x) & \textbf{49.7} \\
\bottomrule
\end{tabular}

\end{minipage}
\hfill
\begin{minipage}{0.40\textwidth}
\centering
\caption{Ablation on tool-call budget.}
\label{tab:tool_calls}

\begin{tabular}{cc|cc}
\toprule
Text & Image & E-VQA & InfoSeek \\
\midrule
2 & 2 & 48.2 & 48.7 \\
3 & 3 & \textbf{52.6} & 53.4 \\
4 & 4 & 52.4 & \textbf{54.3} \\
\bottomrule
\end{tabular}

\end{minipage}

\end{table*}

\subsection{Comparison with State-Of-The-Art Methods}
Table~\ref{tab:main_results} reports the main results on two knowledge-based VQA benchmarks, E-VQA and InfoSeek.
We compare our approach with three representative groups of methods:
(1) \textbf{zero-shot MLLMs} without external knowledge,
(2) \textbf{search-agent methods} that rely on tool interactions, and
(3) \textbf{retrieval-augmented methods} under a fixed RAG-style pipeline.
Overall, our method achieves the best performance across all evaluation metrics on both datasets, and the gains are consistent across different model sizes.

\textbf{Comparison with zero-shot MLLMs.}
Zero-shot MLLMs must answer using only internal knowledge.
However, E-VQA and InfoSeek emphasize long-tail entities and fine-grained attributes that are less likely to be well covered in pretraining.
As a result, relying solely on parametric knowledge leads to low accuracy.

\textbf{Comparison with search-agent methods.}
Our approach also shows clear advantages over existing search-agent baselines.
A key limitation of prior agents is the lack of an explicit correction mechanism: once the first retrieval drifts to a wrong entity, later steps often accumulate evidence around the wrong page, creating a bias that is difficult to correct.

\textbf{Comparison with retrieval-augmented methods.}
We also observe consistent improvements over retrieval-augmented methods.
Existing RAG pipelines often use a fixed retrieval modality and a single-shot retrieval step.
When entity recognition fails, the system cannot correct the retrieved evidence and is forced to answer based on incorrect context.
In addition, some questions require knowledge across multiple Wikipedia pages, which is hard to cover with one retrieval.
In contrast, we formulate KB-VQA as a progressive search-and-reason process.

To further test generalization, we evaluate on OK-VQA (Table~\ref{tab:okvqa_results}).
Our method also achieves clear improvements on OK-VQA, indicating that the learned tool-usage policy generalizes beyond the training benchmarks.

\begin{wrapfigure}{r}{0.6\textwidth}
  \centering
  \vspace{-0.7cm}
   \includegraphics[width=\linewidth]{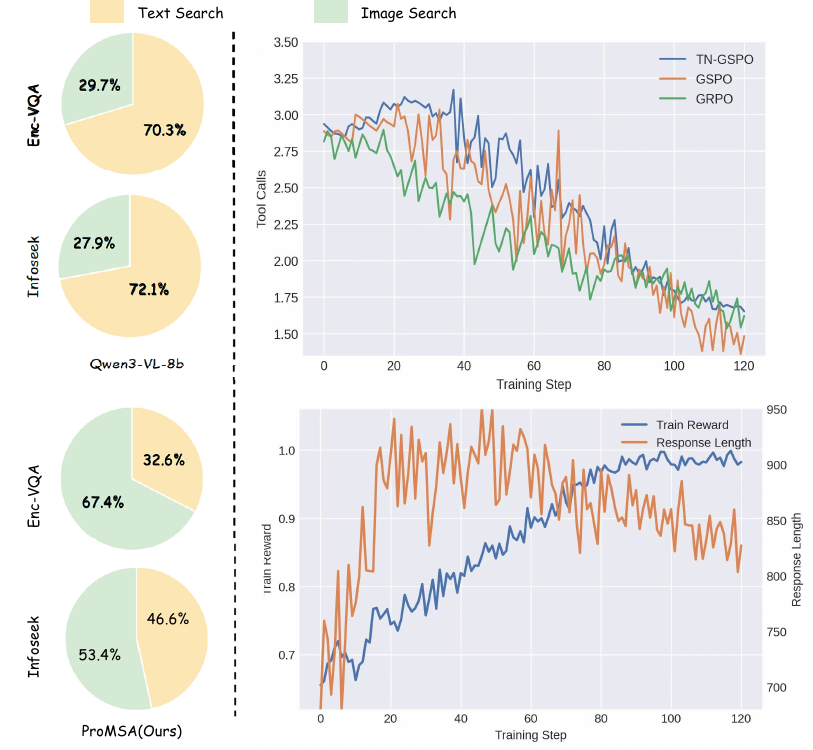}
   \caption{Tool usage and training dynamics of the proposed search agent. Left: proportions of text vs. image search. Right: tool calls, response length, and reward during training across RL strategies.}
   \label{fig:Tool_usage_and_training_dynamics}
\end{wrapfigure}

\subsection{Ablation Study and Visualization}

\textbf{Effect of training stages.}
To quantify the contribution of each training stage, we report results on E-VQA and InfoSeek across three settings (Table~\ref{tab:training_stages}):
(1) \textbf{Base}: our proposed search-agent inference framework without any training;
(2) \textbf{Stage I (Cold Start)}: rejection-sampling SFT on top of Base to learn basic tool-call formats and reasoning/output structure;
(3) \textbf{Stage II (RL Training)}: further reinforcement learning with TN-GSPO.
The Cold Start stage teaches reliable interaction patterns, enabling the model to execute the search process correctly.
The RL stage focuses on learning more effective search strategies over multiple interaction rounds.
With an explicit tool-cost penalty, the agent learns to balance search depth and information gain, leading to substantial improvements.

\begin{table*}[t]
\centering
\scriptsize
\setlength{\tabcolsep}{4pt}
\renewcommand{\arraystretch}{0.95}

\begin{minipage}{0.55\textwidth}
\centering
\caption{Accuracy comparison on the OK-VQA\cite{marino2019ok} dataset.}
\label{tab:okvqa_results}
\begin{tabular}{l c c}
\toprule
\textbf{Method} & \textbf{Model} & \textbf{Accuracy} \\
\midrule
Wiki-PRE-7B & Qwen2.5-VL-7B & 77.8 \\
MMKB-RAG\cite{ling2025mmkb} & LLaMA-3.1-8B & 65.4 \\
CC-VQA & Qwen2.5-VL-7B & 78.8 \\
\midrule
\textbf{Ours} & Qwen2.5-VL-7B & 82.7 \\
\textbf{Ours} & Qwen3-VL-8B & \textbf{85.6} \\
\bottomrule
\end{tabular}
\end{minipage}
\hfill
\begin{minipage}{0.40\textwidth}
\centering
\caption{Ablation on retrieval top-$k$.}
\label{tab:topk}
\begin{tabular}{c|cc}
\toprule
Top-$k$ & E-VQA & InfoSeek \\
\midrule
$k=1$ & 45.9 & 41.7 \\
$k=2$ & 48.3 & 47.6 \\
$k=3$ & \textbf{52.6} & 53.4 \\
$k=4$ & 52.1 & \textbf{54.1} \\
\bottomrule
\end{tabular}
\end{minipage}

\end{table*}

\textbf{Comparison of RL optimization methods.}
Table~\ref{tab:RL_method} compares different RL optimization strategies.
With sequence-level normalization and asymmetric clipping, the policy can explore useful decisions more effectively while maintaining training stability, which yields clear gains.
Furthermore, TN-GSPO extends the normalization by incorporating the tool horizon, so the update scale depends on both generation length and tool interaction depth.
This design supports more stable learning of effective search behaviors and achieves the best performance.

\textbf{Impact of tool availability.}
To study the role of different tools, we evaluate three settings: \textbf{text search only}, \textbf{image search only}, and \textbf{both tools} (Table~\ref{tab:tool_ablation}).
When both text and image search are available, the agent can adaptively choose the appropriate retrieval modality during reasoning, progressively collecting complete evidence and significantly improving final answer accuracy.

\textbf{Inference time analysis.}
We compare the inference time of our method with several existing systems on E-VQA and report the average time per sample (Table~\ref{tab:time_comparison}).
Overall, our method achieves substantially higher accuracy while maintaining competitive inference efficiency, suggesting a good trade-off between performance and computational cost.

\textbf{Training dynamics and tool usage.}
Figure~\ref{fig:Tool_usage_and_training_dynamics} (left) shows the tool usage ratio on E-VQA and InfoSeek.
Before RL, the model tends to rely more on text search.
After RL, the distribution shifts noticeably, indicating that reinforcement learning helps the model learn retrieval behaviors that better match task needs.

Figure~\ref{fig:Tool_usage_and_training_dynamics} (top right) reports the average number of tool calls during training for different sequence-level optimization methods.
Models trained with GRPO reduce tool calls rapidly at an early stage, which suggests the policy may overemphasize tool reduction and suffer from insufficient evidence collection.
In contrast, TN-GSPO gradually stabilizes the tool-call count within a reasonable range, leading to more stable and effective search strategies.

Figure~\ref{fig:Tool_usage_and_training_dynamics} (bottom right) shows the trends of average training reward and response length.
As training proceeds, the average reward increases steadily.
In early training, the model generates longer responses and uses more tool calls to explore strategies.
As the policy improves, it can reach sufficient evidence with shorter reasoning paths, and the response length decreases accordingly.

\begin{figure}[tb]
  \centering
  \includegraphics[width=\linewidth]{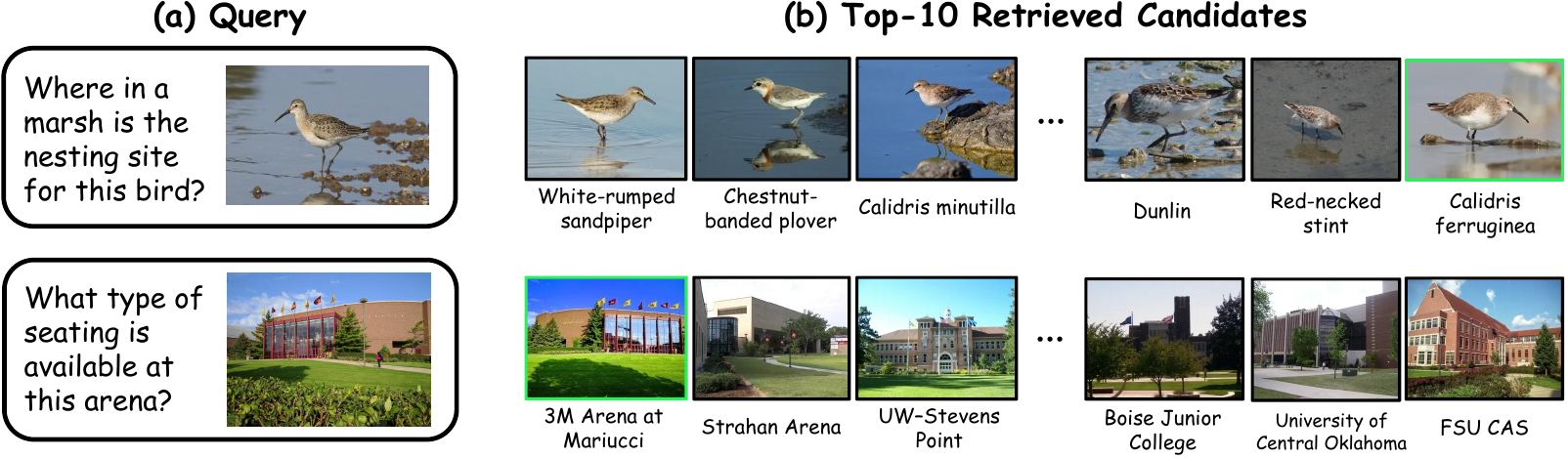}
  \caption{Retrieval examples from the image search module.}
  \label{fig:retrieval}
\end{figure}

\begin{table}[t]
\centering
\footnotesize
\setlength{\tabcolsep}{6pt}
\caption{Retrieval behavior analysis across interaction rounds.}
\label{tab:retrieval_behavior}
\begin{tabular}{lccc}
\toprule
\textbf{Metric} & \textbf{Round-1} & \textbf{Round-2} & \textbf{Round-3} \\
\midrule
Retrieval Acc. & 39.1 & 48.8 & 52.7 \\
Correct Retrieval \& Stop & 32.4 & 45.5 & 50.6 \\
Correct Retrieval \& Continue & 6.7 & 3.3 & 2.1 \\
Incorrect Retrieval \& Continue & 46.4 & 7.4 & 3.2 \\
Incorrect Retrieval \& Stop & 14.5 & 43.8 & 44.1 \\
\bottomrule
\end{tabular}
\end{table}

\textbf{Impact of tool-call budget and retrieval Top-$k$.}
We vary the tool-call budget and retrieval Top-$k$ and evaluate on E-VQA and InfoSeek (Tables~\ref{tab:tool_calls} and~\ref{tab:topk}).
Increasing the tool-call budget or retrieving more content improves the chance of recalling correct evidence, reducing errors caused by failed early retrieval.
However, further increases lead to diminishing returns and can slightly hurt performance.
A likely reason is that larger retrieved sets introduce more noise, which makes evidence compression and reasoning harder and reduces the benefit of additional recall.

\textbf{Retrieval behavior across interaction rounds.}
Table~\ref{tab:retrieval_behavior} provides a fine-grained analysis of retrieval and decision behaviors across interaction rounds for our final model.
We report statistics for samples where a retrieval action is triggered; samples answered directly without retrieval are excluded.
\textbf{Round-1} and \textbf{Round-2} denote behaviors after retrieval in the first and second interaction rounds, respectively.
\textbf{Retrieval Accuracy} measures whether the retrieved knowledge contains at least one correct document.

\textbf{Correct Retrieval \& Stop} is the fraction of cases where the agent stops after retrieving correct evidence.
\textbf{Incorrect Retrieval \& Continue} is the fraction of cases where the agent continues searching after failing to retrieve correct evidence.
All remaining cases (e.g., correct retrieval but continue, or incorrect retrieval but stop) are grouped as \textbf{Others}.

From Round-1, the model tends to continue more often when retrieval is incorrect, showing a strong tendency to correct early failures, while it often stops quickly once key evidence is found.
In Round-2, \textbf{Correct Retrieval \& Stop} further increases to 45.5\%, suggesting that after correction the agent is more likely to converge.
We also observe a relatively high \textbf{Others} ratio in Round-2, mainly because the agent stops more frequently in the second round, and some of these stops happen even when retrieval is still incorrect.
This behavior is influenced by the tool-cost constraint during training.
In addition, for long-tail cases the model may not be fully confident in judging whether the retrieved documents are correct, which can also contribute to this pattern.

\textbf{Retrieval Visualization.}
Figure \ref{fig:retrieval} shows the retrieval visualization.In the first example, the correct entity appears only at rank-10, while in the second it is ranked first. This highlights rank variation and motivates adaptive retrieval.

\section{Conclusion}
We study KB-VQA in the long-tail setting, where models must decide both what to answer and how to search for missing knowledge. 
We propose \textbf{ProMSA}, a progressive multimodal search agent that alternates between image and text retrieval and stops when enough evidence is collected. 
The agent performs retrieval under explicit budgets and avoids repeated results through deduplication. 
We also introduce \textbf{TN-GSPO}, a sequence-level RL objective that normalizes updates by both generation length and tool interaction depth, leading to more stable learning. Experiments on E-VQA and InfoSeek show clear improvements over strong RAG and search-agent baselines.

\bibliographystyle{splncs04}
\bibliography{main}
\end{document}